\documentclass{article}

\usepackage{arxiv}

\usepackage{graphics} 
\usepackage{diagbox}
\usepackage{array,multirow,graphicx}
\usepackage{epsfig} 
\usepackage{mathptmx} 
\usepackage{times} 
\usepackage{amsmath} 
\usepackage{amssymb}  
\usepackage{tikz}
\usepackage{balance}
\usepackage{url}
\usepackage{algorithm}
\usepackage{algorithmicx}
\usepackage[]{algpseudocode}
\usetikzlibrary{bayesnet}
\usetikzlibrary{arrows}
\usepackage{wrapfig}
\usepackage{multirow}
\usepackage{subcaption}

\newbox\MytempboxA
\newbox\MytempboxB

\title{\LARGE \bf
Customized Handling of Unintended Interface Operation \\ in Assistive Robots}

\author{
  Deepak Gopinath\thanks{Equal contribution. Preprint under review.} \\
  Northwestern University, Shirley Ryan AbilityLab \\
  Chicago, Illinois \\
  \texttt{deepakgopinath@u.northwestern.edu} \\
  \And
  Mahdieh Nejati Javaremi$^{*}$ \\
  Northwestern University, Shirley Ryan AbilityLab \\
  Chicago, Illinois \\
  \texttt{m.nejati@u.northwestern.edu} \\
   \And
 Brenna D. Argall\\
 Northwestern University, Shirley Ryan AbilityLab \\
 Chicago, Illinois \\
  \texttt{brenna.argall@northwestern.edu} \\
}
\date{}
\begin{document}
\maketitle

\begin{abstract}
We present an assistance system that reasons about a human's \textit{intended} actions during robot teleoperation in order to provide appropriate corrections for unintended behavior. We model the human's physical interaction with a control interface during robot teleoperation and distinguish between \textit{intended} and \textit{measured} physical actions explicitly. By reasoning over the unobserved intentions using model-based inference techniques our assistive system provides customized corrections on a user's issued commands. We validate our algorithm with a 10-person human subject study in which we evaluate the performance of the proposed assistance paradigms. Our results show that the assistance paradigms helped to significantly reduce task completion time, number of mode switches, cognitive workload and user frustration and improve overall user satisfaction. 
\end{abstract}

\section{INTRODUCTION}
One of the most promising application domains for shared human-robot
control is assistive devices. In this domain, robotics autonomy
collaborates with a human in the operation of their assistive device,
with the aim of increasing that human’s independence and safety. In
addition to its utility as a control signal, the human's input often is used in various other capacities by the robotics autonomy, such as input
to an inference engine. Deviations---in magnitude, direction, or
timing---between the true signal intended by the human and that received
by the autonomy thus can have rippling effects throughout the control
system. Critically, within this domain, the \textit{source} of the
human's signal overwhelmingly is treated as a black box. However, not
only might the human's issued signal be dramatically impacted by their motor
impairment, limitations of the interface accessible to them to operate
the assistive device often mask their true signal intent.
We posit that to consider the source of the human control signal, and
its influence on the correctness of the signal issued, is critical.

When a person is fit for a powered wheelchair, arguably the most
ubiquitous powered assistive device, the seating clinician will choose
the control interface based on the user’s unique physical abilities and
constraints. Operating a control interface requires the human to
\textit{physically activate} the interface---whether via button press, joystick
deflection~\cite{king2012dusty}, screen tap~\cite{cremer2016investigation}, or even electrical signals issued by muscles~\cite{luo2019teleoperation} or
by the brain~\cite{rebsamen2010brain}. The signal which results then is mapped to the control
space of the device. Neither this activation, nor this mapping,
typically is represented within a robotics autonomy system. Instead, the
robot control signal which results is represented in
isolation---independent of the activation or mapping mechanisms.
However, prior work has shown the type of control interface to
significantly effect the timing, transient noise, and accuracy of a signal issued to operate
an assistive device~\cite{nejati2019}. Moreover, when used in partnership with
robotics autonomy, when the autonomy is \textit{not} aware of such
interface usage characteristics as the activation and mapping
mechanisms, the overall performance of the shared-control system
degrades~\cite{nejati2019}. The suggestion is of a need for explicit modeling of the
user’s physical interaction with the interface, and that
\textit{interface-awareness} is a key component to the design of
successful assistive shared-autonomy algorithms.

In this work, we introduce a mathematical model that formalizes the human's physical interaction with the interface and then use this model to provide customized corrections on their issued commands within a shared control framework.

Specifically, our contributions are threefold: 
\paragraph{Modeling the User's Physical Interface Operation} We mathematically formulate the user's \textit{physical interaction} with the control interface during robot teleoperation---specifically how intended user inputs are altered through the interface before being measured by the system. We use data collected from the user to build \textit{user-specific} models of control mapping and stochastic deviations from intended commands to \textit{personalize} the corrective algorithms. 

\paragraph{Model-Based Inference of Intended Input}
Using the interface-aware physical interaction model and some prior knowledge of the user's high-level behavior, we use probabilistic reasoning over latent intended user control commands to deduce unintended deviations from optimal behavior. Notably, our method does not require any additional sensor streams for improving prediction accuracy.

\paragraph{Customized Corrective Assistance} We formulate two methods to provide appropriate modifications to the measured human control input in an \textit{online} fashion. 
The assistance algorithm is personalized to the user because the user-specific probabilistic models encode the idiosyncrasies of a particular user's interaction behavior with the robot. 


\section{RELATED WORK}\label{sec:rw}
Shared control assistive systems often require a good estimate of the user's intent---which could be a high-level goal such as a navigation landmark a user might want to drive a wheelchair towards or an object on a table that a user intends to grab using an assistive robotic arm~\cite{losey2017}. Bayesian inference based approaches are widely used in the context of shared-control in which the user is modelled as a Markov Decision Process and is assumed to be noisily optimizing some cost function with respect to a high-level goal~\cite{admoni2016predicting, javdani2015shared}. In our work, we take a more fine-grained approach to modeling teleoperation in which we explicitly distinguish the intended and measured interface-level physical actions.

Deviations between a person's intended and executed actions can arise due to cognitive as well as physical factors~\cite{ajzen2004}. For motor-impaired people, the inherent physical limitations can increase the likelihood of accidental deviations from intended commands which can lead to unwanted robot behavior. Therefore, in a shared-autonomy system it is important for the autonomy to make decisions based on \textit{intended} as opposed to executed actions to improve the quality of interaction; thus the need for action-level intent inference. Work in driver behavior modeling has investigated action-level inference, for example, to classify and predict driver actions~\cite{pentland1999}. Another work considers the uncertainty in human grasp intent to provide appropriate autonomous robotic grasp plans~\cite{bowman2019}. Unlike our work, they assume the human is physically capable of producing intended commands and the source of uncertainty is due to detection noise. 
Previous work has modeled a person's internal beliefs about a dynamic system, and uses an internal-to-true dynamics transfer function in order to provide the assistance that leads to a desired human action or learning outcome~\cite{reddy2018, rafferty2015}. 
In these works it is assumed that any suboptimal human command is due to a mismatch between their internalized and the true dynamics model and there is no control sharing---the autonomy is always issuing commands based on what it infers to be the user's intent.

\section{MOTIVATION: INTERFACE-AWARE SIGNAL INTERPRETATION}\label{sec:interface_awareness}
We postulate that modeling how the physical actions are mapped to the task-level actions and how the control signal is then altered through the interface will help the autonomy to \textit{reason about the deficiencies} in human teleoperation to \textit{improve the quality} of teleoperation.

\begin{figure}
    \centering
        \begin{subfigure}[b]{.4\textwidth}
            \includegraphics[width=\textwidth]{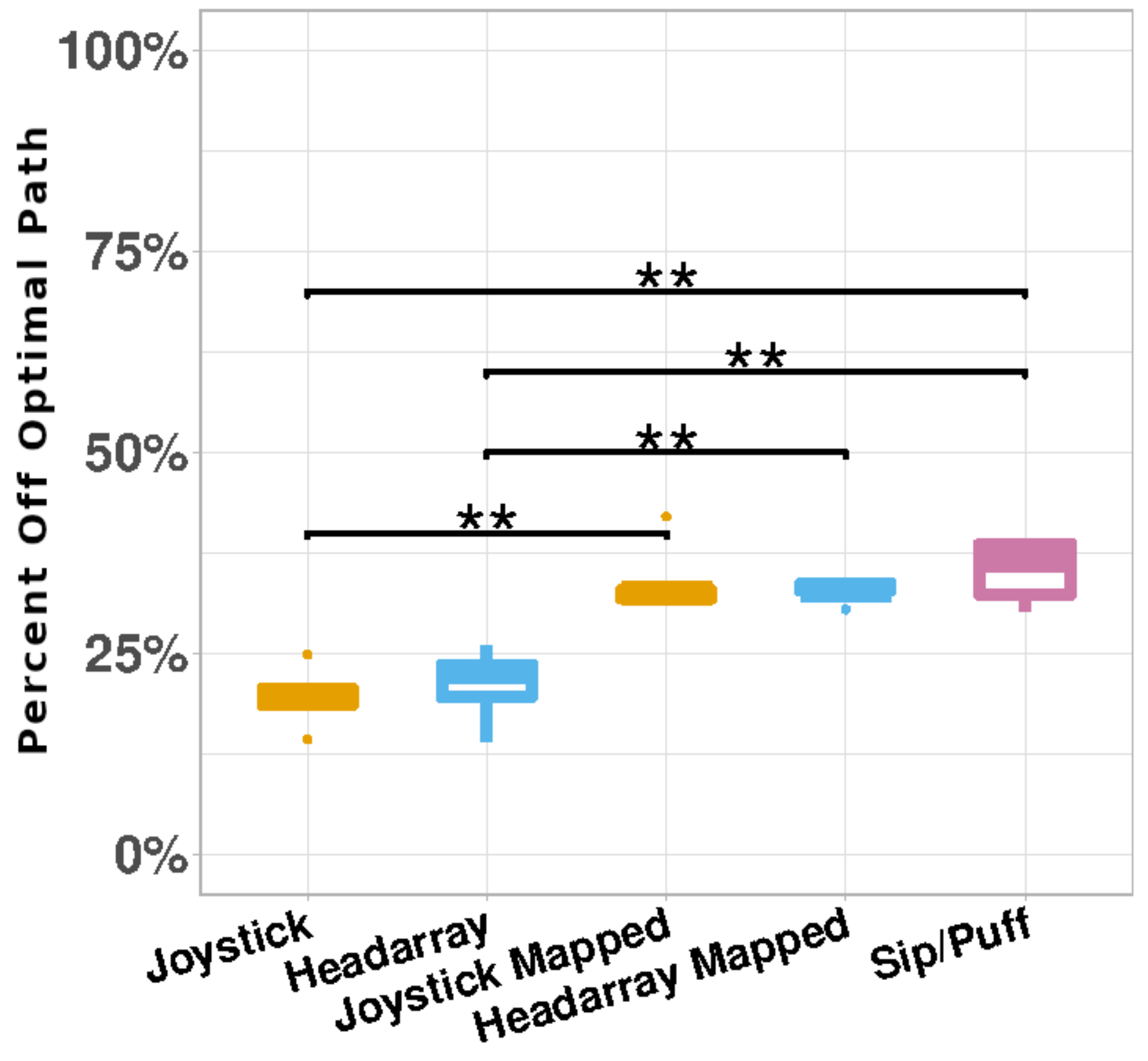}
            \caption{}
            \label{fig:nBB}
        \end{subfigure}
        \begin{subfigure}[b]{.4\textwidth}
            \includegraphics[width=\textwidth]{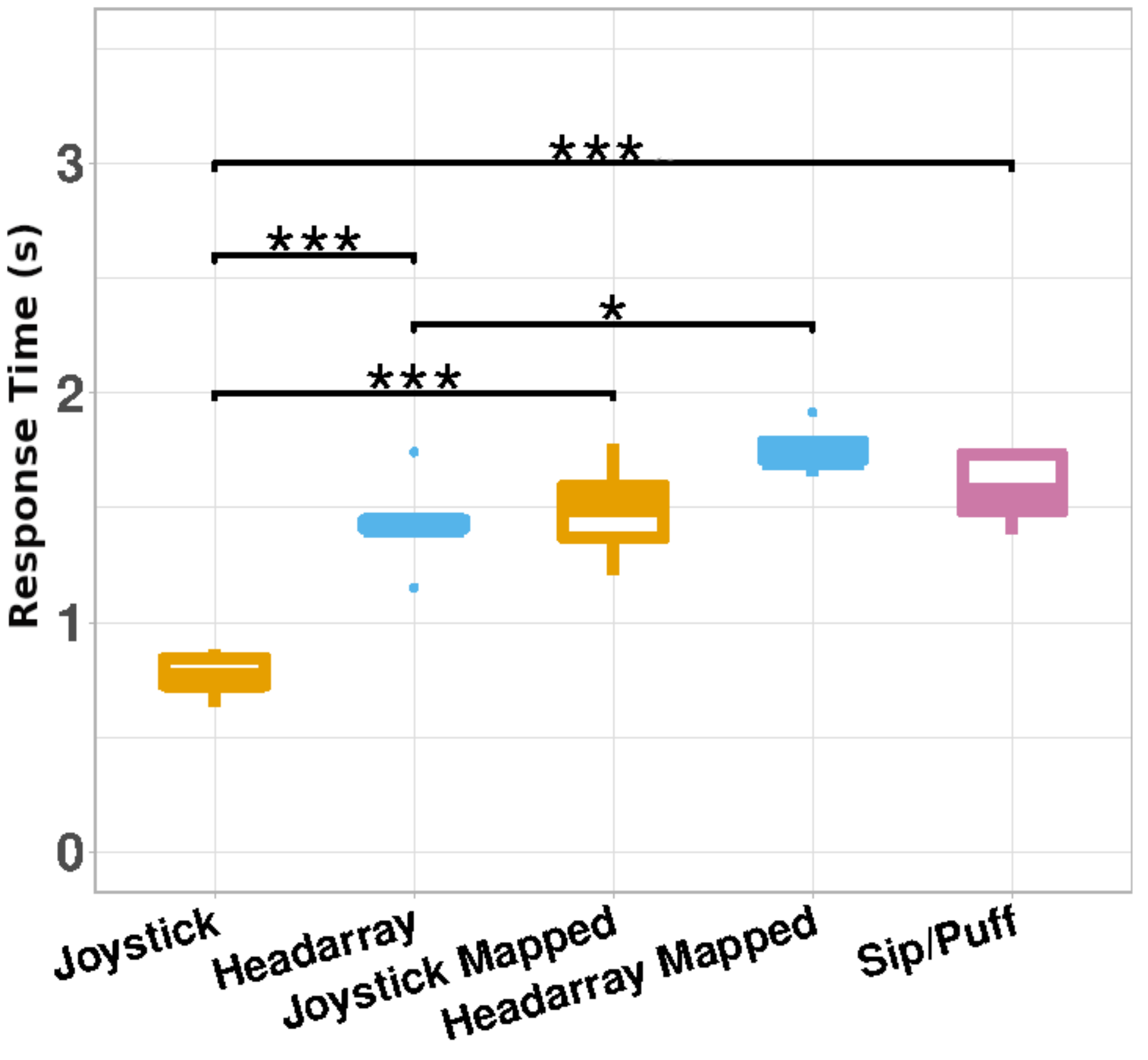}
            \caption{}
            \label{fig:tOB}
        \end{subfigure}
        \caption{(a) Trajectory following performance. (b) Response time to prompted command. The notation ${\ast}$ implies a p-value of $p < 0.05$, ${\ast\ast}$ implies $p < 0.01$, and ${\ast\ast}$${\ast}$ implies $p < 0.001$.}
        \vspace{-0.6cm}
        \label{fig:interface_mapping_comp}
\end{figure}

We investigate the extent to which the physical mechanism and the mapping paradigm of an interface explain the differences in usage characteristics. To that end, we perform a pilot study using three common interfaces employed by powered wheelchair users---namely a joystick, headarray, and sip-n-puff (SNP). We evaluate the operation of these interfaces on two open-source computer game tasks designed to assess trajectory and command following performance~\cite{nejati2019}.\footnote{Source code: \\ \url{https://github.com/argallab/interface_assessments}} To understand the effect of how the signal is altered through the interface, in addition to the most common mappings, we additionally modify the joystick and headarray mapping to be the same as that of the SNP interface. This is the only direction of remapping possible because the joystick and headarray are higher dimensional than the SNP. The results indicate that even though the physical mechanism of providing input is different, when the control mappings are similar, the usage characteristics are normalized across the interfaces (Figure~\ref{fig:interface_mapping_comp}). However, the performance characteristics under joystick and headarray interfaces suffer with the new mapping which
motivates the need for an interface-aware autonomous assistance system that will compensate for the degradation in overall human-robot team performance.

Our approach to interface-aware signal interpretation is to explicitly model both how the physical actions are mapped to the task-level actions through the interface and how the user signal is stochastically altered through the interface.

\section{MATHEMATICAL FORMALISM}\label{sec:formalism}
In this section we describe our mathematical model of the user's physical interaction with a control interface during manual teleoperation and the assistive algorithm which uses our model to provide customized assistance. 
\subsection{Modeling the User's Physical Interface Operation}\label{ssec:user_modelling}
\begin{figure}
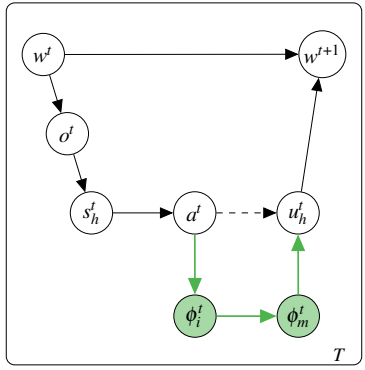

\vspace{0.3cm}
\centering
\scalebox{0.8}{
    \tikz{
    \definecolor{c}{rgb}{0.3, 0.7, 0.3}
 \node[latent] (wt) {$w^t$};%
 \node[latent, below=of wt, yshift=0.4cm, xshift=0.4cm] (ot) {$o^t$};
 \node[latent, below=of ot, yshift=0.4cm, xshift=0.4cm] (Sht) {$s_h^t$}; 
 \node[latent, right=of Sht] (at) {$a^t$};
 \node[latent, right=of at] (uht) {$u_h^t$};
 \node[obs, fill=c!50, below=of at, xshift=0.1] (kit) {$\phi_i^t$};
 \node[obs, fill=c!50, right=of kit] (kmt) {$\phi_m^t$};
 \node[latent, right=of wt, above=of kmt, yshift=2.6cm, xshift=.4cm] (wtp1) {$w^{t\text{+}1}$};
 
\plate [inner sep=.25cm,yshift=.2cm] {plate1} {(wt)(ot)(Sht)(at)(uht)(kit)(kmt)(wtp1)} {$T$}; %
\edge{wt}{ot}
\edge{ot}{Sht}
\edge{Sht}{at}
\edge{uht}{wtp1}
\edge{wt}{wtp1}
\draw [dashed, ->](at) -- (uht);
\draw [c, thick, ->](at) -- (kit);
\draw [c, thick, ->](kit) -- (kmt);
\draw [c, thick, ->](kmt) -- (uht);
 }
 }
\caption{Probabilistic graphical model depicting a specific user's interactions with the robot via the control interface at single time step $t$. The dashed edge between $a^t$ and $u^t$ indicates how teleoperation is typically modeled. The nodes and edges that model the physical aspect of controlling the interface are highlighted in green.} 
\label{fig:pgm_model}
\vspace{-0.5cm}
\end{figure}

Figure~\ref{fig:pgm_model} depicts the probabilistic graphical model of a user's physical interaction with a control interface during robot teleoperation at a single time step $t$.

Let $w^t$ represent the true world state and $o^t$ the partial observation of the world state. 
$s_h^t \in \mathcal{S}$ denotes the human's internal state  and encodes the human's goals and beliefs. $a^t$ represents the action primitives that are defined in the task space that the user intends to execute at time step $t$. $u_h^t$ is the low-level control commands issued to the robot. $\phi_i^t \in \Phi$ is the \textit{intended interface-level physical action} initiated by the user that aims to achieve $a^t$ and is \textit{unobserved}. The set of available interface-level physical actions depends on the physical modality used for activating an interface. $\phi_m^t \in \Phi$ is the \textit{measured interface-level physical action} produced by the user and is fully \textit{observed}. The novel contribution of this model is in a) the explicit modeling of the interface-dependent physical mechanisms that generate $u_h^t$ and b) in distinguishing the latent $\phi_i^t$ from the measured $\phi_m^t$. In a noise-free setting, $\phi_i^t$ and $\phi_m^t$ are equivalent. However, in practice, $\phi_m^t$ may deviate from $\phi_i^t$ due to biases resulting from motor-impairment, stress or equipment malfunction, to name a few. 


\subsection{Estimation of $a^t$ from measured $\phi_m^t$}\label{ssec:estimation_framework}
We are interested in the following question: given the measured interface-level physical action issued by the user $\phi^t_m$ what is the probability distribution over the task-level action primitives $a^t$? More precisely, we are interested in the probability distribution $p \big( a^t | \phi_m^t \big)$. Concretely, using Bayes theorem, we have

\begin{equation}\label{eq:1}
    p\big(a^t  |\phi_m^t\big) \; \propto p\big(\phi_m^t | a^t\big)p\big(a^t\big)
\end{equation}

Marginalizing over $\phi_i^t$ we have,
\begin{equation}\label{eq:2}
    p\big(\phi_m^t  | a^t\big) = \Sigma_{\phi^t_i \in \Phi} p\big(\phi_m^t, \phi_i^t  | a^t\big)
\end{equation}

Due to the conditional independence of $a^t$ and $\phi_m^t$ Equation~\ref{eq:2} becomes
\begin{equation}\label{eq:3}
    p\big(\phi_m^t  | a^t\big) = \Sigma_{\phi^t_i \in \Phi} p\big(\phi_m^t| \phi_i^t\big)p\big( \phi_i^t | a^t\big)
\end{equation}

and plugging Equation~\ref{eq:3} in Equation~\ref{eq:1} we have,
\begin{equation}\label{eq:4}
    p\big(a^t  |\phi_m^t\big) = \eta p\big(a^t\big) \Sigma_{\phi^t_i \in \Phi} p\big(\phi_m^t| \phi_i^t\big)p\big( \phi_i^t | a^t\big)
\end{equation}
where $\eta$ is the normalization factor. 
We also have 
\begin{equation}\label{eq:5}
    p\big(a^t\big) = \Sigma_{s_h^t \in \mathcal{S}} p\big(a^t| s_h^t\big)p\big( s_h^t\big)
\end{equation}

and combining Equation~\ref{eq:5} with Equation~\ref{eq:4} we have 

\begin{equation}\label{eq:6}
    p\big(a^t  |\phi_m^t\big) =  \eta\Sigma_{s_h^t \in \mathcal{S}} p\big(a^t| s_h^t\big)p\big( s_h^t\big)\Bigg[\Sigma_{\phi^t_i \in \Phi} p\big(\phi_m^t| \phi_i^t\big)p\big( \phi_i^t | a^t\big)\Bigg] .
\end{equation}
Each one of the three conditional probability distributions that appear in the right hand side of Equation~\ref{eq:6} have intuitive interpretations. 
$p\big(a^t| s_h^t\big)$ is the \textit{control policy} the user follows during task execution. With training, practice, and learning, the user's policy will gradually converge to an optimum---with respect to an internal cost function~\cite{ito2000, jagacinski1978, kluzik2008}. 
$p\big(\phi_i^t | a^t\big)$ captures the user's \textit{internal model} of the mapping from task-level action primitives to the intended interface-level physical actions. Users acquire an internal model of this mapping (which is static and deterministic) via training~\cite{pierella2019}. 
Finally, $p\big(\phi_m^t | \phi_i^t\big)$ captures the stochastic deviations of the \textit{measured} interface-level physical actions from the \textit{intended} interface-level physical actions and is the \textit{user input distortion model}. These conditional probability distributions can be personalized by learning from user-specific teleoperation data.

\vspace{-0.1cm}
\subsection{Customized Handling of Unintended Physical Actions}

The motivation for our framework described in Section~\ref{ssec:estimation_framework} is to improve the control of complex robotic machines with limited interfaces used by people with motor-impairments. Equation~\ref{eq:6} can be used within a shared-control assistive paradigm to infer the human's true task-level intent and provide corrective measures, if necessary, to reduce the cognitive and physical burden of unintentional interface operation. The inference scheme is outlined in Algorithm~\ref{alg:inference}. Using Equation~\ref{eq:6}, at every time step $t$ we compute the likelihood of $a^t \in \mathcal{A}$ conditioned on the observed $\phi^t_m$ (line 2). The action primitive corresponding to the maximum of the distribution is inferred as the intended action $a^t_{inferred}$, and using the true control mapping function $f$ we compute $\phi^t_{inferred}$ (lines 3-4). In Algorithm~\ref{alg:shared-control}, the autonomy intervenes only if $\phi^t_{inferred}$ is different from $\phi^t_m$ and the uncertainty of prediction, computed as the entropy $H$ of the distribution, is less than a predefined threshold $\epsilon$. Otherwise, $\phi^t_m$ will be passed through the pipeline unimpeded.
The appealing characteristic of our proposed control-sharing algorithm is that the user is maximally in control which potentially can improve user satisfaction and acceptance~\cite{broad2019}. Most notably, our assistance system remains as close to the manual system as possible and does not rely on augmenting the human-robot team with high fidelity sensors to improve prediction accuracy. This is important for reasons of user adoption and cost. Furthermore, when the autonomy steps in, it does so only to provide commands closest to the user's true intentions (which they were unable to issue correctly themselves in the first place).

We implement and evaluate two corrective assistance shared-control paradigms.  

\begin{algorithm}[t]

\vspace{0.2cm}
    \caption{Infer Intended Commands}
    \label{alg:inference}
    \begin{algorithmic}[1]
        \Procedure{infer\_intended\_command}{$t, \phi_m^t$}
            \State{compute $p\big(a^t | \phi_m^t\big)$}\Comment{equation~\ref{eq:6}}   \State{$a^t_{inferred} \xleftarrow[]{} argmax\big((p\big(a^t|\phi^t_m\big)\big)$}
            \State{$\phi^t_{inferred} \xleftarrow[]{} f\big(a^t_{inferred}\big)$}\Comment{true control mapping}
            \State \Return{$\phi^t_{inferred}$}
        \EndProcedure{}
    \end{algorithmic}
\end{algorithm}
\begin{algorithm}[t]
    \caption{Handle Unintended Commands}
    \label{alg:shared-control}
    \begin{algorithmic}[1]
        \Procedure{handle\_unintended\_commands}{$t, \phi_m^t$}
            \State $\phi^t_{inferred}$ =  \Call{infer\_intended\_command}{$t, \phi_m^t$}
            \If{$\phi^t_{inferred} \ne \phi^t_m $}
                \If{$H\big(p\big(a^t | \phi^t_m\big)\big) < \epsilon$}\Comment{uncertainty is low}
                    \If{filter\_based\_assistance}
                        \State{$\phi^t_{corrected}=0$}
                    \ElsIf{corrective\_based\_assistance}
                        \State{$\phi^t_{corrected}=\phi^t_{inferred}$}
                    \EndIf
                \Else
                    \State \Return{$\phi_m^t$}
                \EndIf
            \Else
                \State\Return{$\phi_m^t$}
            \EndIf
        \State \Return{$\phi_{corrected}^t$}
        \EndProcedure{}
    \end{algorithmic}
\end{algorithm}
\subsubsection{Filtering autonomy}
Conditioned on the uncertainty in the prediction, if $\phi^t_m$ is deemed as unintended, filter (or block) this command. In our implementation this means that $\phi^t_{corrected}=0$ (i.e. no motion or mode-switching occurs as a result of this command). 

\subsubsection{Corrective autonomy}
Conditioned on the uncertainty in the prediction, if $\phi^t_m$ is deemed as unintended, $\phi^t_{corrected} = \phi^t_{inferred}$ (i.e. the control command that will result in the inferred intended action). 

\section{SIMULATION-BASED ALGORITHM EVALUATION}\label{sec:sim_study}
In order to gain a deeper insight into how different hyper-parameters---such as noise levels in $p\big(\phi^t_i | a^t\big)$ and $p\big(\phi^t_m | \phi^t_i\big)$---affect the overall performance of our proposed assistance algorithm, we designed a simulation-based experiment. We chose a path-navigation task for this simulation-based evaluation (shown in Figure~\ref{fig:sim_study_env}) and assumed that an SNP interface was being used for robot teleoperation. The domain of task-level and interface-level actions for an SNP are defined in Section~\ref{ssec:exp_testbed}. Task-level action primitives ($a^t$) and intended interface-level physical actions ($\phi_i^t)$ were sampled from the generative model shown in Figure~\ref{fig:pgm_model}. $\phi_i^t$ was corrupted according to $p(\phi_m^t | \phi_i^t)$ to generate $\phi_m^t$.


\begin{figure}[t]
\vspace{0.2cm}
    \centering
         \begin{subfigure}[b]{.235\textwidth}
            \includegraphics[width=\textwidth]{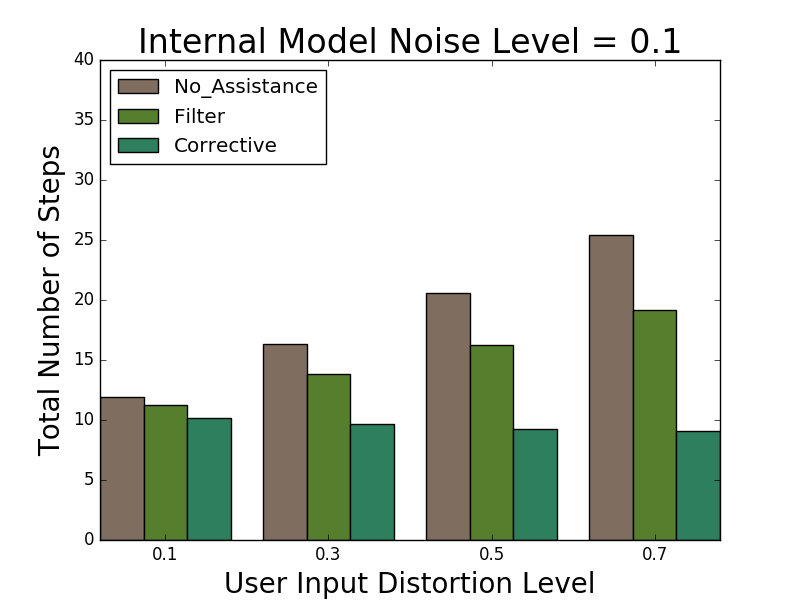}
            \caption{}
            \label{fig:total_num_steps_01}
        \end{subfigure}
        \begin{subfigure}[b]{.235\textwidth}
            \includegraphics[width=\textwidth]{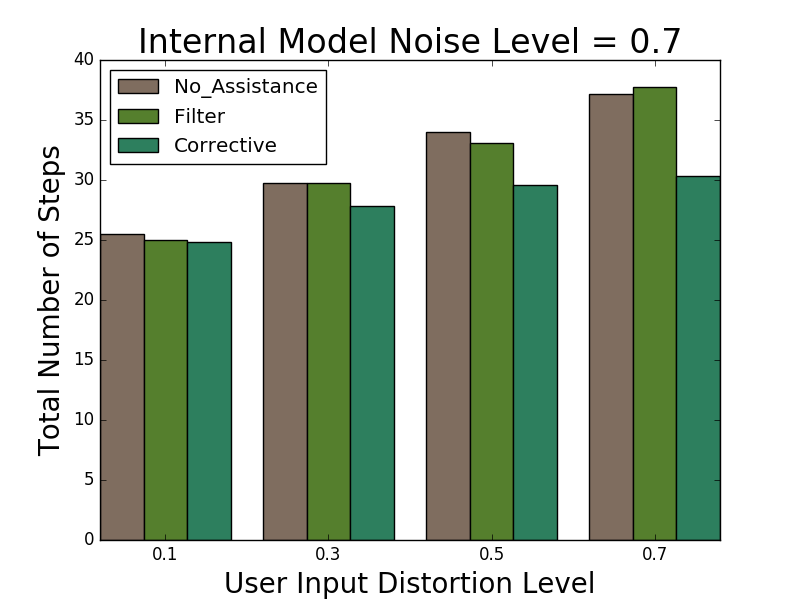}
            \caption{}
            \label{fig:total_num_steps_07}
        \end{subfigure}
    \caption{Total number of state transitions for two different noise levels in $p(\phi_i | a)$ - 0.1 (left) 0.7 (right). }
    \label{fig:sim_total_num_steps}
\end{figure}
In our simulations, the amount of random noise injected into $p(\phi_i^t | a^t)$ and $p(\phi_m^t | \phi_i^t)$ was treated as a scalar simulation parameter.\footnote{0.0 and 1.0 denote a delta and uniform random distribution, respectively.} Table~\ref{tbl:sim_params} indicates the ranges of all parameters used in the simulation experiments.
\begin{table}[h!]
    \centering
    \begin{tabular}{|p{3cm}|p{4cm}|}
    \hline
        Parameter & Range of Values \\ \hline \hline
        $N$ - Number of Turns & [1,2,3] \\ \hline
        Assistance Type & [Filter, Corrective, No Assistance] \\ \hline
        Noise in $p(\phi_i | a)$ & [0.1, 0.3, 0.5, 0.7] \\ \hline
        Noise in $p(\phi_m | \phi_i)$ & [0.1, 0.3, 0.5, 0.7] \\ \hline
    \end{tabular}
    \caption{Ranges of different simulation parameters.}
    \label{tbl:sim_params}
    \vspace{-0.5cm}
\end{table}

We evaluated the performance of our assistance algorithm as measured by the total number of state transitions during a trial under different assistance conditions.

Figure~\ref{fig:sim_total_num_steps} reveals that a more accurate internal model (where $p(\phi_i^t | a^t)$ has low corruption noise), in general, will help the user perform better. For a given $p(\phi_i^t | a^t)$, the corrective assistance paradigm has the highest performance followed by filtering and no-assistance. The difference in performance between the assistance paradigms decreases as the noise in $p(\phi_i^t | a^t)$ increases, illustrating the need for proper training and acquisition of accurate internal models. These insights guided our experimental design which will be explained in detail in the next section.\footnote{For additional simulation-based evaluation please refer to the supplementary video.}

\section{EXPERIMENTAL DESIGN}\label{sec:exp_design}

We ran a human-subject study ($n=10$) to evaluate our inference algorithm and assistive paradigms in terms of overall task performance and user preference. All participants gave their informed, signed consent to participate in the experiment which was approved by Northwestern University's Institutional Review Board. 
Each study session consisted of three phases. \underline{Phase 1:} Training and data collection to model $p\big(\phi^t_i | a^t\big)$, \underline{Phase 2:} Training and data collection to model $p\big(\phi^t_m| \phi_i^t\big)$, and \underline{Phase 3:} Assistance evaluation phase in which the subjects controlled a 3D point robot using the SNP interface under three distinct assistance conditions.


\subsection{Experimental Testbed} \label{ssec:exp_testbed}
For the evaluation task we designed a simulated navigation environment (Figure~\ref{fig:sim_study_env}).\footnote{Code available at \\ \url{https://github.com/argallab/customized_interface_aware_assistance.git}} The evaluation tasks were designed in a simulated environment specifically because it allowed us to develop a three dimensional control task requiring a greater number of mode switches, thus making it more difficult compared to a non-holonomic powered wheelchair. 

We used the 1-DoF SNP as our control interface as it is one of the most difficult to use, and is often the only device accessible to those with severe motor impairments. The subjects used the SNP to operate a 3D point robot's motion along $x$, $y$ or $\theta$ dimension, one at a time.\footnote{The dimensionality mismatch between the interface and the robot necessitates the control space to be partitioned into smaller subsets called \textit{modes}. Motion is restricted only along those dimensions that belong to the currently active mode. The user can use the interface to activate different modes by switching between them and this is referred to as \textit{mode switching}.} The set $\mathcal{A}$ of task-level action primitives consisted of (1) clockwise mode switch (2) counter-clockwise mode switch 3) positive direction motion and 4) negative direction motion. The set ${\Phi}$ of interface-level physical actions available for a sip-and-puff were 1) hard sip 2) soft sip 3) hard puff and 4) soft puff. The true correspondence between $a^t$ and $\phi_i^t$ was deterministic (denoted as $f(\cdot)$) and predefined. When using an SNP the likelihood of generating unintended control actions was quite high because of (1) same method of input used for both motion control and mode switching (2) inherent difficulty in breath regulation and (3) factors such as fatigue, stress, and saliva gathered in the straw. 


\subsection{Learning Personalized Distributions}
We designed two tasks to capture the personalized distributions $p\big(\phi_i^t | a^t\big)$ and  $p\big(\phi_m^t | \phi_i^t\big)$ from user studies. 

\subsubsection{Personalized Internal Control-Mapping Model}\label{sss:control_mapping}
\begin{figure}[t!]
    \centering
    \includegraphics[width=0.5\textwidth]{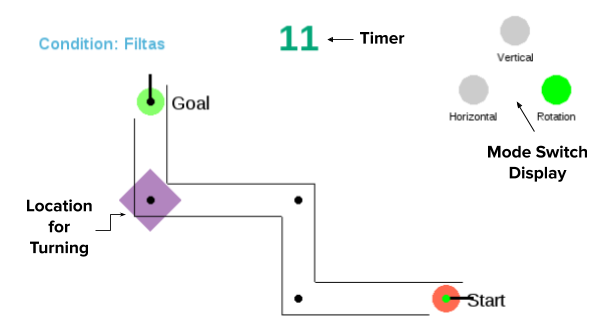}
    \caption{An example trial in the navigation environment used in our human-subject study. Each trial had a maximum time limit of 50 seconds. Feedback regarding the current active mode was displayed on the screen (on the top right corner).}
    \label{fig:sim_study_env}
    \vspace{-0.5cm}
\end{figure}

Participants were first trained on the true mapping ($f(\cdot)$) during a standardized training phase. The training phase consisted of three phases: (1) learning about the action primitives ($a^t$) for the 3D experimental task-space (Figure~\ref{fig:sim_study_env}), (2) learning about the interface-level physical actions ($\phi^t$) available through the interface, and (3) the mapping between $\phi^t$ and $a^t$. 
The training was followed by six blocks of testing trials. During testing, the user was shown a graphical depiction of $a^t$, and instructed to select the correct $\phi^t$. Each testing trial had a time limit of five seconds. 
The subjects repeated the training and the testing protocol until they met a minimum level of proficiency. 
The distribution $p\big(\phi_i^t | a^t\big)$ was modeled using data collected during the testing phase.

\subsubsection{Personalized User Input Distortion Model}
In order to model $p\big(\phi_m^t | \phi_i^t\big)$, the user was trained on the operation of the interface in order to ensure good understanding of physical aspects of using the interface.
We designed a second task to model $p\big(\phi_m^t | \phi_i^t\big)$.
During testing, the user was shown a command on the screen (e.g. ``Soft Puff") and asked to issue the same command through the interface. Each trial had a time limit of five seconds. 
The distribution $p(\phi_m^t | \phi_i^t)$ was modeled using the data collected during this testing phase. 

\subsection{Assistance Evaluation}


In this task, the subject controlled the motion of a 3-DoF point robot along predefined paths from a start pose to a goal pose. The optimal number of mode switches to complete the task was known. For each trial, the start and end positions were randomized. The initial mode was random, but different from the optimal mode the user had to be in for performing the first movement. 
Users performed the evaluation task under three conditions: (1) \textit{no assistance}, (2) \textit{filtering} assistance, and (3) \textit{corrective} assistance. The subject was required to rotate the point robot to the target orientation at one of the corners (highlighted in violet). Subjects were prompted to complete the task efficiently (i.e. least number of mode switches and in a timely manner). A trial was deemed successful if the robot was at the goal pose within the allotted time limit. Subjects performed six blocks (two blocks per assistance condition) of six trials each. In total, we collected 360 trials (120 trials per assistance condition). After each block, the subjects were required to respond to a NASA-TLX questionnaire. At the end of the final block, the subjects filled out a post-session survey in which they rank ordered the different assistance conditions according to their preference, intuitiveness, helpfulness, and difficulty. 

\section {RESULTS}\label{sec:results}

We analyze group performances using the non-parametric Kruskal-Wallis test and perform the Conover's test post-hoc pairwise comparisons to find the strength of significance.\footnote{For all figures in Section~\ref{sec:results}, the notation ${\ast}$ implies a p-value of $p < 0.05$, ${\ast\ast}$ implies $p < 0.01$, and ${\ast\ast}$${\ast}$ implies $p < 0.001$.}

\subsection{Objective Task Performance Metrics}

\begin{figure}[t!]
    \centering
         \begin{subfigure}[b]{.235\textwidth}
            \includegraphics[width=\textwidth]{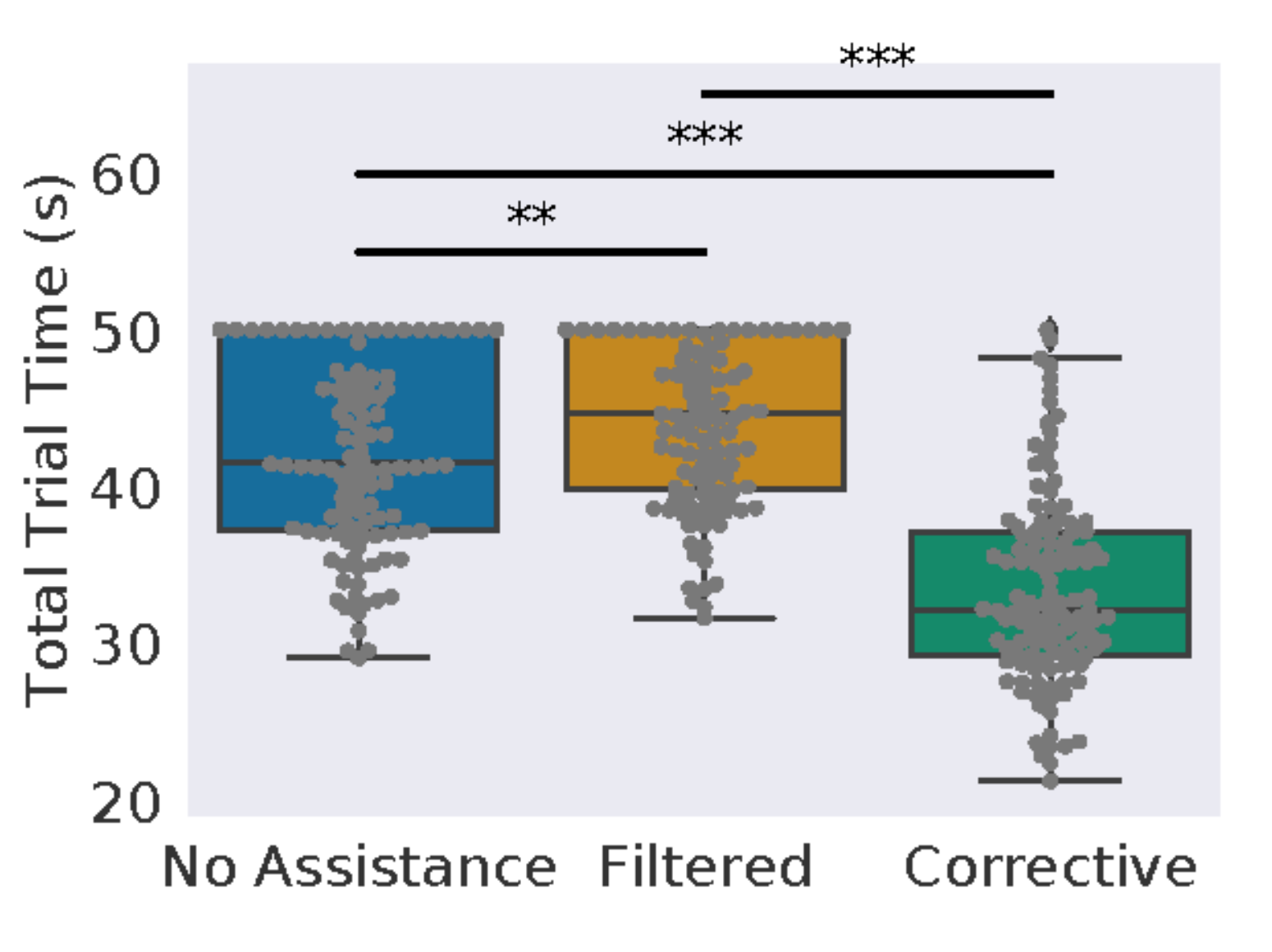}
            \caption{}
            \label{fig:time}
        \end{subfigure}
        \begin{subfigure}[b]{.235\textwidth}
            \includegraphics[width=\textwidth]{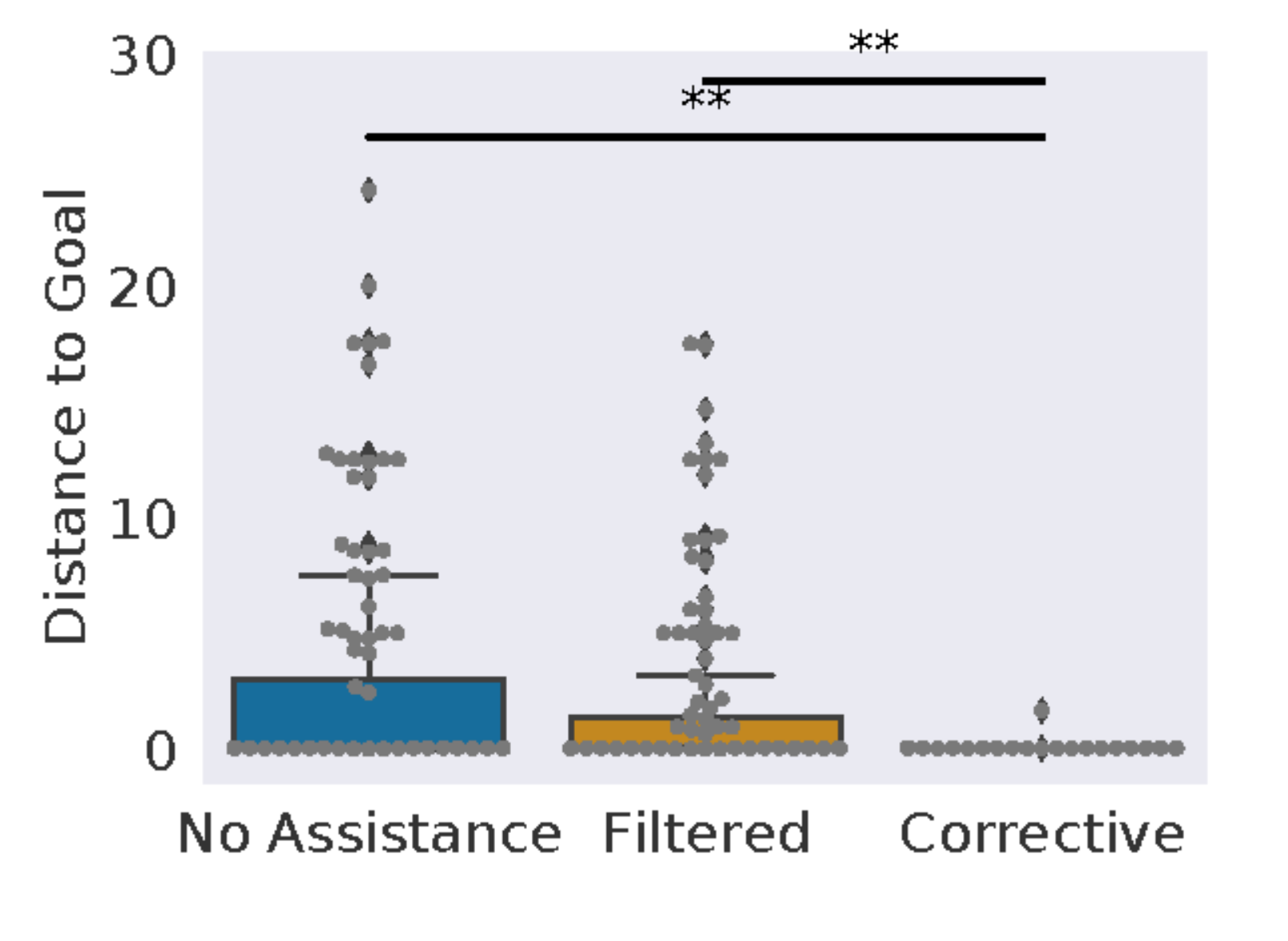}
            \caption{}
            \label{fig:distance}
        \end{subfigure}
        \begin{subfigure}[b]{.235\textwidth}
            \includegraphics[width=\textwidth]{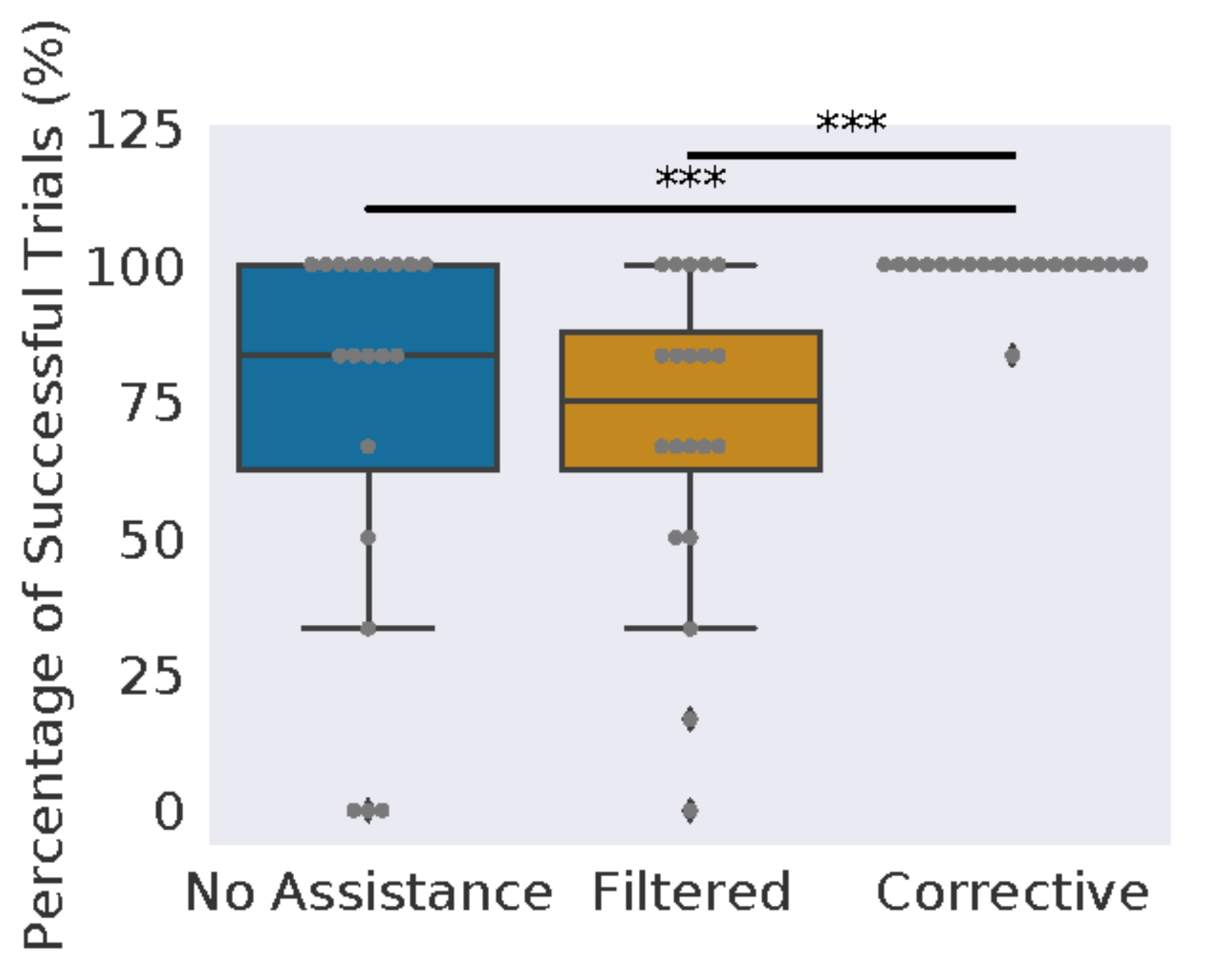}
            \caption{}
            \label{fig:success}
        \end{subfigure}
        \begin{subfigure}[b]{.235\textwidth}
            \includegraphics[width=\textwidth]{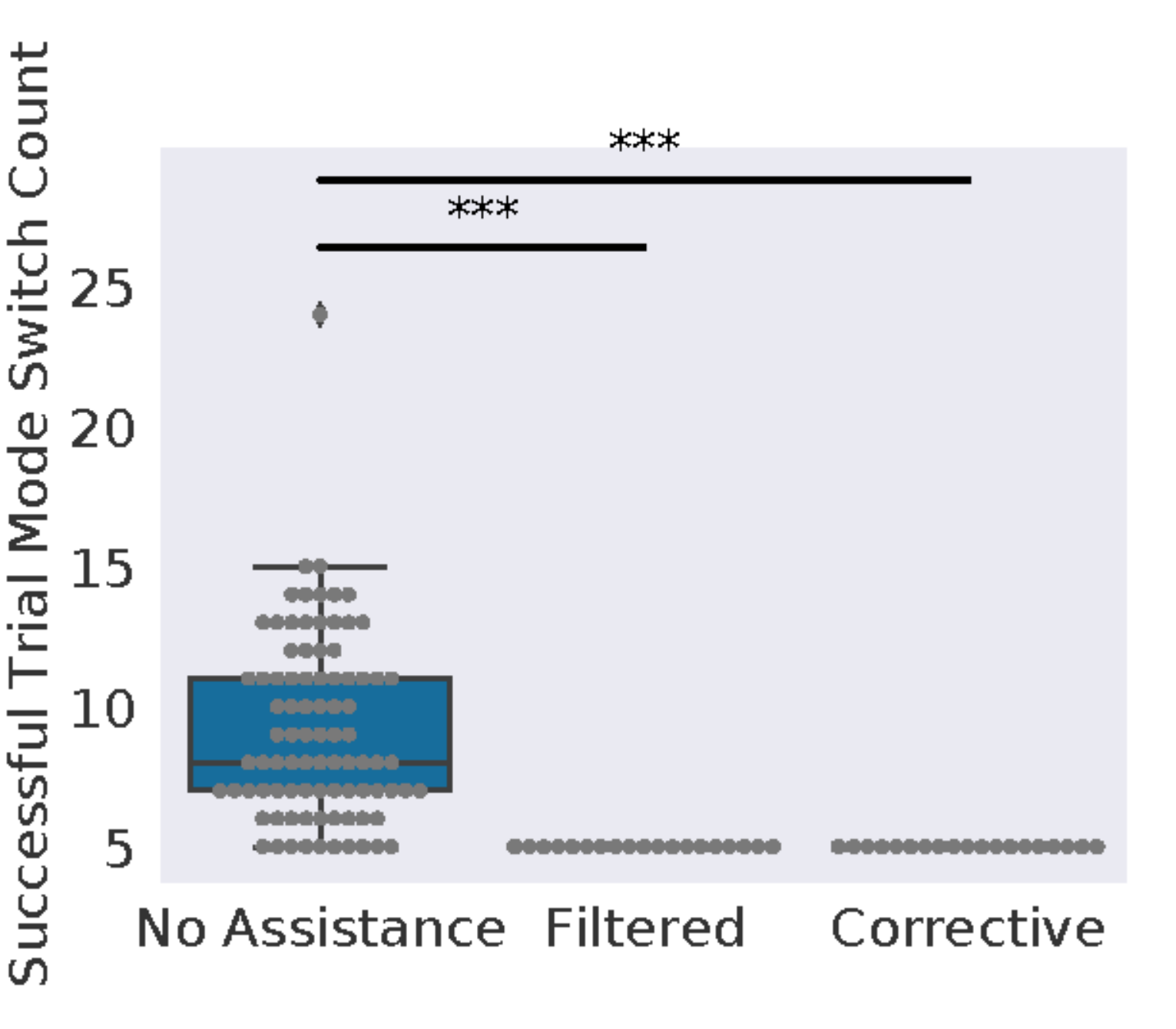}
            \caption{}
            \label{fig:mode_switches}
        \end{subfigure}
        \caption{Objective task performance metrics grouped by assistance condition. (a) Total trial time with maximum trial time capped at 50s, (b) distance to the goal at the end of trials, (c), percentage of successful trials, (c) total number of mode switches during successful trials. All metrics improve significantly with the corrective assistance condition.}
        \vspace{-.8cm}
        \label{fig:performance}
\end{figure}

To evaluate the effectiveness of our algorithm on overall task performance, we compare (1) the total task completion times, (2) the distance to the desired goal position and orientation at the end of each trial, (3) the percentage of successful trials under each assistance condition, and (4) the total number of mode switches for successful trials, across the three assistance conditions (Figure~\ref{fig:performance}). 

As seen in Figure~\ref{fig:time}, the total trial time is shortest under \textit{corrective} assistance, increases with the \textit{no assistance}, and is largest under the \textit{filtered} assistance paradigm. In the \textit{filtered} assistance condition, repeated issuance of sub-optimal commands is always blocked due to the strict enforcement of optimal mode switches. This results in increased task completion time because the user may need to think about what the optimal input is. Contrarily, under the \textit{no assistance} condition, the subject could end up in the desired control mode by executing multiple consecutive sub-optimal mode switches in quick succession. Specifically, in our experimental setup, two counter-clockwise mode switches is equivalent to a single clockwise mode switch and vice versa. 
\begin{wrapfigure}{r}{0.25\textwidth}
\vspace{-0.5cm}
\centering
\includegraphics[height=3.1cm]{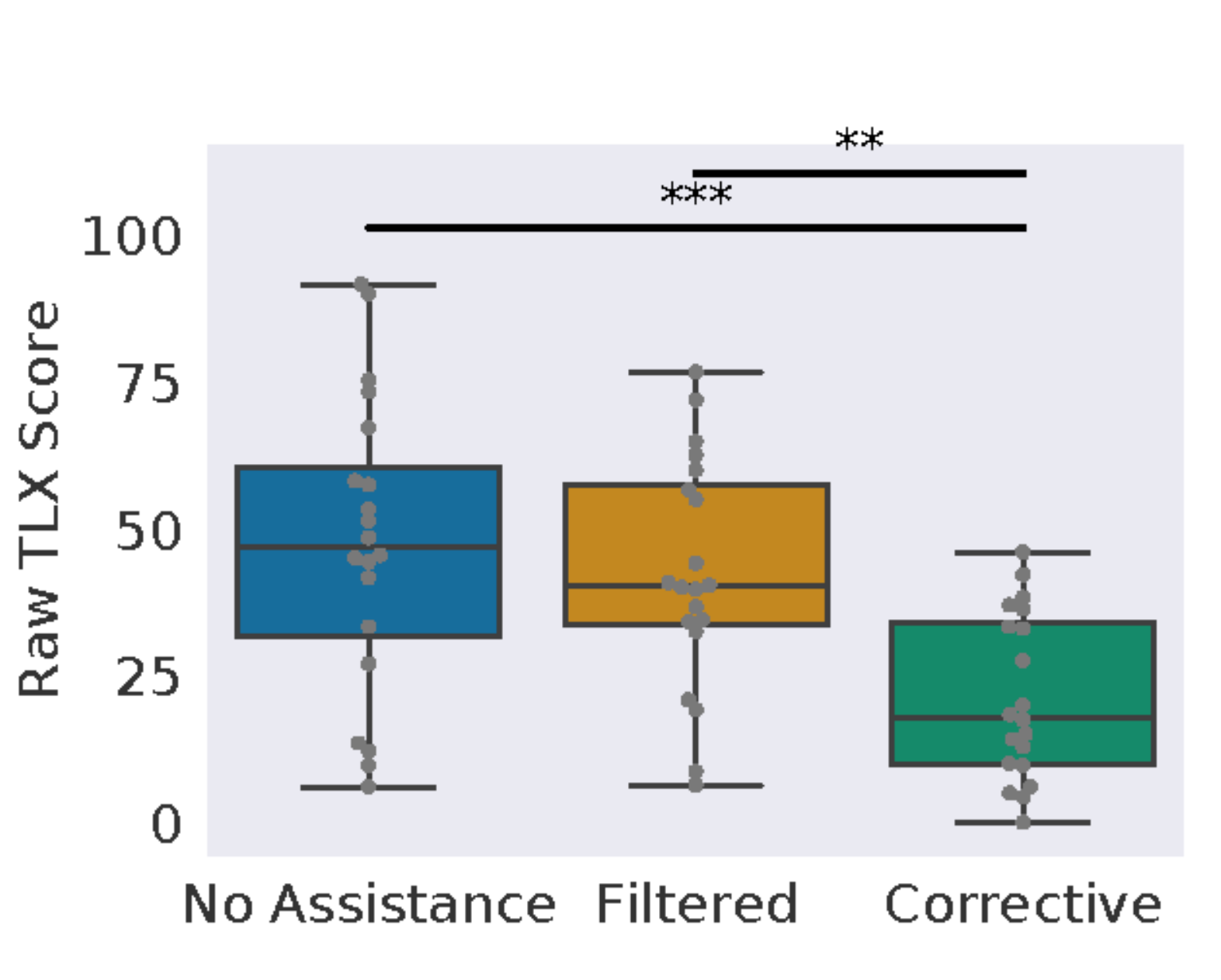}
\caption{Perceived workload measured by the NASA-TLX score, grouped by assistance condition. }
\label{fig:tlx}
\end{wrapfigure}

Figures~\ref{fig:distance}-\ref{fig:success} show the distance to goal at the end of the trial and the percentage of trials successfully finished by each subject, respectively. Both of these metrics improve significantly under the \textit{corrective} assistance condition. 

The \textit{filtered} and \textit{corrective} assistance paradigms are comparable when looking at the total number of mode switches during successful trials (Figure~\ref{fig:mode_switches}). Both assistance conditions are optimal with respect to the number of mode switches, which is five for all trials. Under \textit{no} assistance, despite successful task completion, the number of executed mode switches is up to three times the optimal number required.  

\subsection{Subjective Task Performance Metric}

We use the raw NASA-TLX as a subjective measure of perceived workload~\cite{tlx}. 
A larger TLX score indicates a higher perceived workload. During corrective assistance, the autonomy offloads some of the cognitive burden by correcting unintended actions and as a result reduces the physical burden of corrections without the user's awareness, as evident by the significant reduction of the user's perceived workload (Figure~\ref{fig:tlx}). 
Although during filtered assistance, the autonomy gives feedback to the user by way of blocking unintended actions, the user is still fully responsible for all issued commands.

\subsection{User Acceptance of Assistive Autonomy}

\begin{figure}[t]
    \centering
        \includegraphics[width=\textwidth]{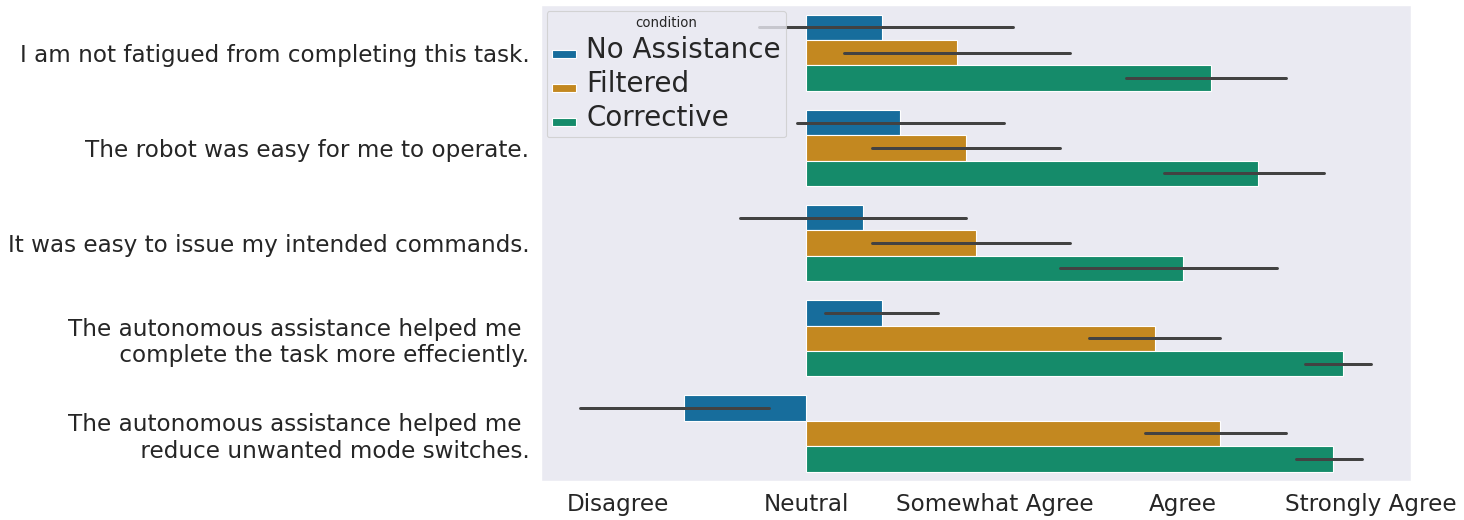}
    \caption{Average user response to post-task questionnaire. The bars indicate standard deviation. }
    \label{fig:ranking}
\end{figure}

We evaluate user preferences and acceptance of our shared-control assistive paradigms using a questionnaire (Figure.~\ref{fig:ranking}). The statements are rated on a 7-point Likert scale from strongly disagree (1) to strongly agree (7). Although some of the objective measures of task performance between \textit{filtered} assistance and \textit{no} assistance are comparable, the users feel that the \textit{filtered} assistance helps them complete the task more efficiently and is easier to operate than under \textit{no assistance}. Overall, the participants show a strong preference toward the \textit{corrective} assistance. 

\section{DISCUSSION AND IMPLICATIONS}\label{sec:discussion}
We have presented a probabilistic graphical model that distinguishes between intended versus measured user control interface signals. We introduced two assistance
paradigms that reason about stochastic deviations in user input when providing corrective behaviors. The efficacy of the assistance paradigms were evaluated both in simulation and via human subject study, with results demonstrating improvements in objective task metrics as well as user perception metrics. 

Our study results suggest that the assistance framework can potentially help users improve their interface operation skills. In particular, since the \textit{filtering} assistance paradigm blocks all user inputs that do not correspond to optimal actions, a user operating the interface under this condition can learn to issue the correct commands, thereby improving overall teleoperation skill. This paradigm might be used within a rehabilitative setting with various methods of feedback as a training paradigm, for example. By contrast, the \textit{corrective} assistance paradigm might help users who are not yet highly skilled or who have plateaued in their skill in operating the interface. A person with a recent motor impairment could begin operating a powered wheelchair quicker with corrective assistance while going through rehab and training, for example, accelerating their mobility independence. 

Each of the assistive paradigms thus have unique advantages that are crucial when operating an assistive device. Concisely, the corrective assistance paradigm is more effective in situations where efficient and successful task completion is more critical (for example, crossing a street on a powered wheelchair), whereas the filtering paradigm can help in maximal skill acquisition. When used in tandem with an adaptive shared-control framework, the proposed assistance paradigms have the potential to improve the overall quality
of device operation while also encouraging skill development and making independent operation of the device more accessible to user.

\section*{ACKNOWLEDGMENT}
This  material  is  based  upon  work  supported  by  the  National  Science  Foundation  under Grant  No.  1552706 and Grant CNS 1544741. Any opinions, findings, and conclusions or recommendations expressed  in  this  material  are  those  of  the  authors and  do not  necessarily  reflect  the  views  of  the  National  Science Foundation.

\bibliographystyle{unsrt}

\balance

\end{document}